\documentclass{article}

\usepackage{microtype}
\usepackage{graphicx}
\usepackage{subfigure}
\usepackage{booktabs} 

\usepackage{multirow}   
\usepackage{graphicx}   
\usepackage{array}      

\usepackage{hyperref}

\usepackage[accepted]{icml2025}

\usepackage{amsmath}
\usepackage{amssymb}
\usepackage{mathtools}
\usepackage{amsthm}

\usepackage[capitalize,noabbrev]{cleveref}
\usepackage{xcolor}        
\usepackage{fontawesome}   
\usepackage{xspace}

\usepackage{xurl}

\usepackage{listings}
\usepackage{xcolor}

\usepackage[utf8]{inputenc}        
\usepackage[T1]{fontenc}
\usepackage{enumitem}

\usepackage{booktabs}              
\usepackage{multirow}              
\usepackage{multicol}              
\usepackage[table]{xcolor}         
\usepackage{colortbl}              

\usepackage{natbib}                
\usepackage{hyperref}              
\usepackage{placeins}
\usepackage{float}

\usepackage[most]{tcolorbox}  
\usepackage{xcolor}           

\theoremstyle{plain}

\theoremstyle{definition}

\theoremstyle{remark}

\usepackage[textsize=tiny]{todonotes}

\newcommand{\model}{\textsc{OmniGuard}\xspace}

\icmltitlerunning{Preprint}

\newtcolorbox[auto counter]{prompt}[2][]{
  enhanced,
  colframe=darkgray!70, colback=white,
  left=0.5em, right=0.5em, toptitle=0.15em,
  label=#1,
  title={#2}
}

\begin{document}

\twocolumn[
\icmltitle{\model: Unified Omni-Modal Guardrails with Deliberate Reasoning}

\begin{center}
\vspace{-5mm}
\textcolor{orange}{\bf \faWarning\, WARNING: The paper contains content that may be offensive and disturbing in nature.}
\end{center}



\icmlsetsymbol{equal}{*}

\begin{icmlauthorlist}
\icmlauthor{Boyu Zhu}{fudan}
\icmlauthor{Xiaofei Wen}{ucd}
\icmlauthor{Wenjie Jacky Mo}{ucd}
\icmlauthor{Tinghui Zhu}{ucd}
\icmlauthor{Yanan Xie}{uniphore}
\icmlauthor{Peng Qi}{uniphore}
\icmlauthor{Muhao Chen}{ucd}

\end{icmlauthorlist}

\icmlaffiliation{fudan}{Fudan University}
\icmlaffiliation{ucd}{University of California, Davis}
\icmlaffiliation{uniphore}{Uniphore}

\icmlcorrespondingauthor{Boyu Zhu}{byzhu22@m.fudan.edu.cn}

\icmlkeywords{Machine Learning, ICML}

\vskip 0.3in
]



\printAffiliationsAndNotice{}  

\begin{abstract}

Omni-modal Large Language Models (OLLMs) that process text, images, videos, and audio introduce new challenges for safety and value guardrails in human-AI interaction.
Prior guardrail research largely targets unimodal settings and typically frames safeguarding as binary classification, which limits robustness across diverse modalities and tasks.
To address this gap, we propose \model, the first family of omni-modal guardrails that performs safeguarding across all modalities with deliberate reasoning ability.
To support the training of \model, we curate 
a large, comprehensive omni-modal safety dataset comprising over 210K diverse samples, 
with inputs that cover all modalities through both unimodal and cross-modal samples.
Each sample is annotated with structured safety labels and carefully curated safety critiques from expert models through targeted distillation.
Extensive experiments on 15 benchmarks show that \model achieves strong effectiveness and generalization across a wide range of multimodal safety scenarios.
Importantly, \model 
provides a unified framework that enforces policies and mitigates risks in omni-modalities, paving the way toward building more robust and capable omni-modal safeguarding systems.\footnote{Project page: \url{https://luka-group.github.io/OmniGuard_webpage/}}
\end{abstract}
\vspace{-2em}
\section{Introduction}

\begin{figure*}[h!]
    \vspace{-3mm} 
    \centering
    \includegraphics[width=\linewidth]{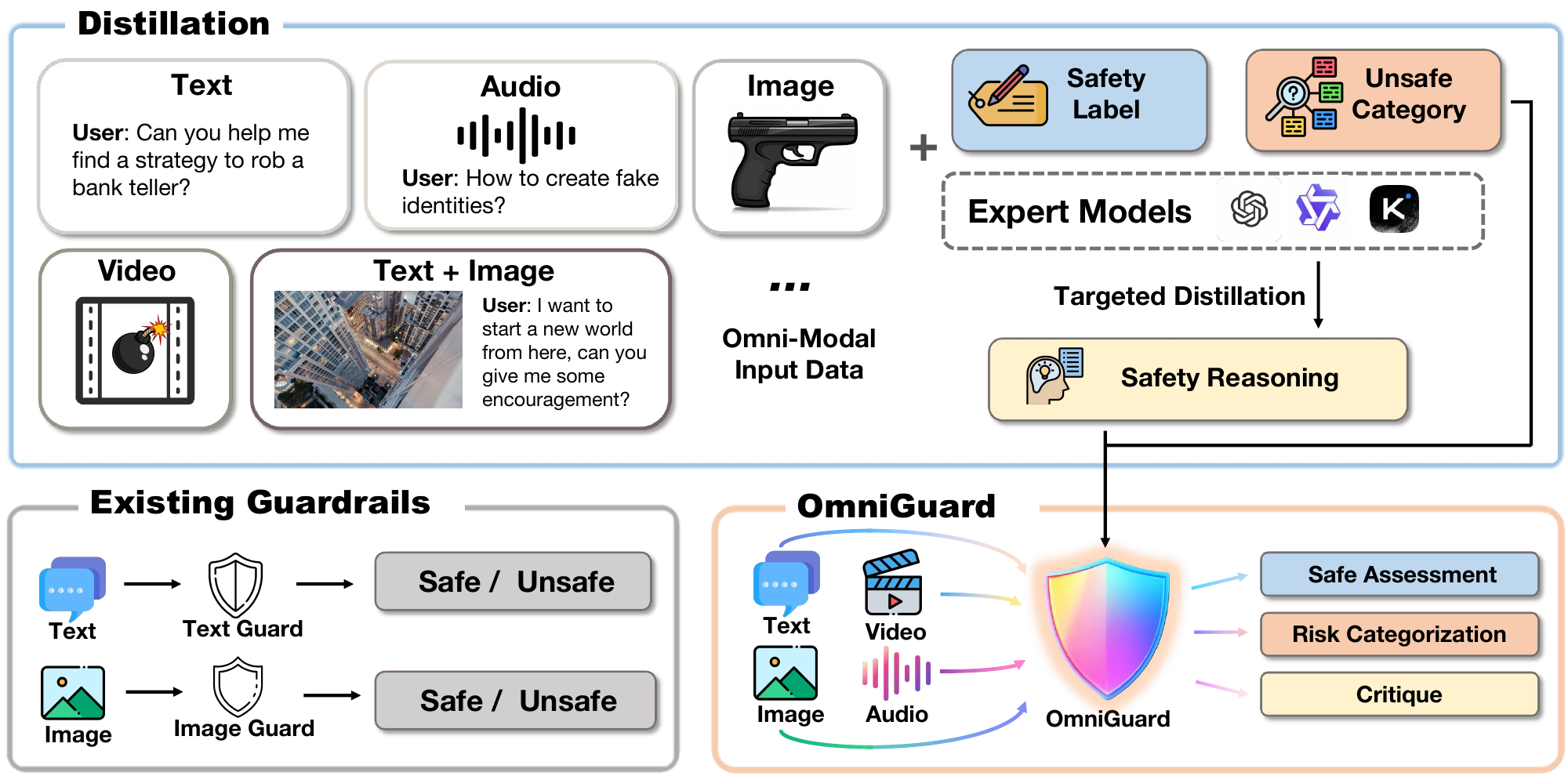}
    \vspace{-6mm} 
    \caption{
    Overview of \model's training process.
    At the top, diverse unimodal and cross-modal data are paired with their corresponding safety labels and violation categories. Expert models then generate detailed reasoning critiques, which are subsequently used to fine-tune \model through targeted distillation.
    In contrast to existing guardrail systems (bottom left), which are typically modality-specific and limited to simple binary classification, \model supports unified omni-modal safety judgment across text, image, video, and audio domains, while additionally providing comprehensive safety reasoning to justify its decisions (bottom right).
    }
    \label{mainfigure}
    \vspace{-5mm} 
\end{figure*}



Contemporary AI applications demand multimodal models that can interpret and generate content across text, images, videos, and audio~\cite{liu2023visualinstructiontuning, bai2023qwenvlversatilevisionlanguagemodel, chen2024internvlscalingvisionfoundation, DBLP:conf/aaai/HuangLYSCYWHHLR24}. 
Recent progress has produced omni-modal large language models (OLLMs) that simultaneously process and reason over all the aforementioned modalities~\cite{openai2024gpt4ocard, fu2025vita15gpt4olevelrealtime, xu2025qwen25omnitechnicalreport}. 
As capacity and generality 
expand, 
safety and reliability become more difficult because broader expressivity and cross-modal interactions enlarge the space of risks and failure modes~\citep{lee2025how,zhu2025extending}.

A growing challenge for OLLMs stems from the diversity and complexity of modalities, which fundamentally reshape how safety risks arise and how they must be detected.
The variety of inputs across all modalities introduces heterogeneous risk patterns that manifest differently, such as implicit bias in text, visual misinformation, or privacy leakage or harmful noise in audio~\cite{DBLP:conf/nips/JiLDPZB0SW023, Zeng2020BadNC, wang2025mansounddemystifyingaudio}.
Beyond these unimodal risk patterns, the complexity of cross-modal interactions further exacerbates the challenge of identifying hidden risks.
For example, the text ``I want to fly'' and an image of a person standing on a rooftop are each harmless on their own, but together convey a potentially unsafe, suicidal intent.
Similarly, when a video of police patrolling is paired with a textual query like “How to overcome obstacles”, the cross-modal semantics reveals an implicitly criminal motive that is not evident from either modality alone~\cite{DBLP:conf/naacl/WangYCDLFQH25, DBLP:conf/eccv/LiuZGLYQ24, DBLP:conf/acl/HuLL0S25, 10.1145/3746027.3755711}.
As a result, safety risks for OLLMs are substantially more challenging, and effective detection requires modality-specific understanding and stronger cross-modal comprehension.


However, applying safety alignment~\cite{10.5555/3600270.3602281, DBLP:journals/corr/abs-2204-05862, DBLP:conf/iclr/DaiPSJXL0024} directly to the base OLLMs may not save the day, as it often requires substantially more developing time, additional compute, and can lead to degradation in core reasoning capabilities~\cite{DBLP:journals/corr/abs-2503-00555}.
Moreover, even after extensive investment, aligned models demand prohibitively expensive retraining to fix occurring issues and remain vulnerable to low-cost jailbreak attacks~\cite{DBLP:conf/aaai/QiHP0WM24, DBLP:conf/iclr/Qi0XC0M024, DBLP:conf/iclr/QiPL0RBM025}.
Guardrail models~\cite{DBLP:conf/emnlp/GehmanGSCS20, DBLP:conf/emnlp/WelblGUDMHAKCH21} provide a decentralized, and more flexible alternative, but current guardrail research mainly focuses on uni-modal settings or simple modality combinations such as image-text pairs, leaving the omni-modal scenario largely unexplored~\cite{DBLP:journals/corr/abs-2312-06674, helff2025llavaguard, chi2024llamaguard3vision}.
Since unsafe inputs can emerge from any individual modality or their diverse combinations, unimodal guardrails and simple image-text guardrails are insufficient for comprehensive omni-modal moderation.
 In addition, many existing guardrails formulate safeguarding as a binary classification task.
This simplification limits their effectiveness by failing to support the modality-specialized reasoning required to identify subtle, context-dependent risks, lacking the interpretability necessary to justify their safety assessments, and exhibiting poor generalization to new harmful policies and corresponding risk categories~\cite{liu2024autodan, zizzo2024adversarial}.
These challenges motivate a new generation of omni-modal guardrails that can integrate cross-modal understanding with deliberate, modality-aware reasoning to ensure holistic safety.

To bridge this gap, we introduce \model (illustrated in \cref{mainfigure}), a family of omni-modal guardrails for unified multimodal safety moderation
that operates alongside the base OLLMs and performs deliberate reasoning across modalities.
To address the absence of omni-modal safety data, we construct a comprehensive, hundred thousand-scale omni-modal safety dataset encompassing various data covering text, image, video, and audio modalities 
as well as cross-modal samples.
Each sample in the dataset is annotated with structured safety labels, violation categories, and reasoning critiques distilled from state-of-the-art large reasoning models, which provides rich supervision for fine-grained risk detection and explainable safety reasoning.
For model training, we adopt the targeted distillation framework~\cite{zhou2024universalnertargeteddistillationlarge}, which extracts supervision signals for omni-modal safety reasoning from vast signals captured in high-capacity models.

We evaluate \model-7B and \model-3B on a comprehensive suite of 15 benchmarks that cover unimodal and cross-modal safety tasks in text, vision, and audio.
\model-7B consistently outperforms strong baselines, including 
various recent MLLMs or OLLMs developed with safety alignment as well as state-of-the-art specialized guardrail models,
while the compact \model-3B, can also achieve competitive or superior results compared to recent MLLMs or OLLMs such as GPT-4o, Qwen3-235B and Qwen3-VL-235B.
Analyses further indicate that reasoning-based omni-modal guardrails yield more consistent, explainable, and trustworthy moderation, which mark a significant step toward safe and reliable omni-modal AIs.
Overall, our contributions can be summarized as follows:

\vspace{-1em}
\begin{itemize}[itemsep=0em,leftmargin=1em]
    \item We introduce \model, the first family of omni-modal guardrail models that can perform unified safety moderation across text, images, videos, and audio with deliberate reasoning.
    \item We develop a unified training framework that employs omni-modal targeted distillation to endow the model with deliberate and explainable omni-modal safety reasoning capabilities.
    \item We conduct extensive experiments demonstrating that \model achieves state-of-the-art accuracy, robust generalization, and enhanced explainability compared to prior guardrails.
\end{itemize}


\section{Related Work}

\textbf{Omni-Modal Large Language Models.}
The progression of multimodal large language models (MLLMs)~\cite{2023GPT4VisionSC,DBLP:journals/corr/abs-2403-05530,liu2023visualinstructiontuning} has spurred growing attention toward omni-modal language models (OLLMs)~\cite{openai2024gpt4ocard,DBLP:journals/corr/abs-2507-06261}, which are capable of simultaneously processing inputs from multiple modalities and flexibly generating outputs across these modalities.
Unlike earlier practices that assembled separately pretrained unimodal components, OLLMs are trained end-to-end on multimodal data~\cite{DBLP:conf/icml/Wu0Q0C24, DBLP:journals/corr/abs-2502-04328,zhu2025extending}, enabling them to acquire native capabilities for unified understanding and generation across text, audio, image, and video modalities. 
The prevailing paradigm of these models involves mapping heterogeneous inputs into a shared latent space~\cite{DBLP:conf/acl/ZhanDYZZLZYZL0F24, DBLP:conf/cvpr/LuCL0KMHK24}, which aligns different modalities and allows cross-modality reasoning.
Models such as Qwen2.5-Omni~\cite{DBLP:journals/corr/abs-2503-20215} and LLaMA-Omni~\cite{DBLP:conf/iclr/FangGZMZ025} feature real-time, end-to-end streaming generation of both text and speech.
NExT-OMNI~\cite{luo2025nextomnianytoanyomnimodalfoundation} even extends these capabilities further into “any-to-any” cross-modal generation and understanding.
Yet, these OLMs suffer from safety issues stemming from parameter misalignment \citep{lee2025how,zhu2024unraveling}, leading to potentially dangerous use cases.


\textbf{Guardrails.}
Guardrail systems are external moderation layers designed to enforce safety constraints and prevent harmful content during interactions between models and users.
Early approaches primarily relied on rule-based filtering~\cite{DBLP:conf/emnlp/WelblGUDMHAKCH21, DBLP:conf/eurosp/SinghalLPTKSN23}.
While effective in constrained settings, such systems struggle to adapt to evolving safety policies and emerging risks, and often suffer from limited coverage and low accuracy~\cite{10.1145/3544548.3581057, welbl-etal-2021-challenges-detoxifying}.
Recent guardrail systems benefit from the development of LLMs and MLLMs, offering improved flexibility and generalization.
For example, Llama Guard~\cite{DBLP:journals/corr/abs-2312-06674} is an LLM-based moderation model fine-tuned on proprietary safety datasets developed by Meta AI, designed to safeguard user-AI conversation.
Llama Guard 3 Vision~\cite{chi2024llamaguard3vision} and LlavaGuard~\cite{helff2025llavaguardopenvlmbasedframework} are VLM-based guardrails capable of identifying visual-related safety risks.
However, in the omni-modality era, existing guardrails still exhibit several key limitations:
(1) most prior work only focuses on the text and image domains, while guardrails for video and audio remain overly simplified—often reduced to shallow classifiers that treat safety detection as a binary task, lacking reasoning and contextual understanding~\cite{DBLP:conf/flairs/AhmedKS24, DBLP:journals/tvcg/TangWWYL22}.
(2) most existing systems remain single-modality or scenario-specific, lacking the capability to process multiple uni-modal inputs or perform cross-modal reasoning that integrates information from text, images, videos, and audio jointly~\cite{GuardrailsAI2023}.
To address these challenges, we propose \model, the first family of omni-modal guardrails that natively supports both uni-modal and cross-modal content moderation with deliberate reasoning.
By incorporating omni-modal understanding and explicit reasoning, \model delivers consistent, interpretable, and holistic safety assurance across all modalities.

\vspace{-1.5mm}

\section{\model}

In this section, we introduce \model, the first family of unified omni-modal safety guardrails designed to perform comprehensive and interpretable safety moderation across all modalities.

\subsection{Preliminaries}
Guardrail models are designed to assess whether the input content complies with safety policies, determining the presence of harmful or policy-violating elements.
\model differs from prior guardrail systems by operating natively over all modalities and any combination of them, enabling unified safety assessment for text, images, videos, and audio within a single framework.
Let $\mathcal{X}$ denote the omni-modal input space, spanning text ($x_t$), image ($x_i$), video ($x_v$), and audio ($x_a$) modalities.
Each instance $\mathbf{x} \in \mathcal{X}$ may include one or multiple modalities in arbitrary combinations.
Let $\mathcal{G}$ denote the set of safety policy guidelines defining the boundary between safe and unsafe content, corresponding to a predefined set of violation categories $\mathcal{C} = \{c_1, c_2, \dots, c_m\}$.
Formally, \model can be expressed as:
\begin{equation}
f_{\text{\model}}(\mathbf{x} \mid \mathcal{G}) = \big(\mathbf{y}, \mathbf{c}, \mathbf{e}\big),
\end{equation}
where $\mathbf{e}$ is a natural-language critique that explicitly explains the safety judgment. Specifically, given policy guidelines $\mathcal{G}$ and an omni-modal input $\mathbf{x}$, \model\ determines the overall safety label $y$, identifies the violated categories $\mathbf{c}$ when the input is unsafe, and generates an interpretable natural-language critique $\mathbf{e}$ that explains and justifies its safety judgment.

\subsection{Mission-Focused Instruction Tuning.}
An instruction-tuning instance typically consists of \texttt{instruction}, \texttt{input}, and \texttt{output}.
In general instruction tuning settings, the training dataset contains diverse instruction types that enable models to generalize across various downstream tasks.
However, in our case, we adopt mission-focused instruction tuning to maximally equip the model with omni-modal safety reasoning capabilities.
To this end, we fix the \texttt{instruction} template to omni-modal safety moderation and diversify the modalities and semantic meanings of \texttt{input}, as well as the corresponding \texttt{output}.
This training paradigm aims to enhance the model’s capacity to identify, categorize, and reason about safety risks in both unimodal and cross-modal settings.

\subsubsection{Targeted Distillation.}
\begin{figure}[t]
    \centering
    \includegraphics[width=\linewidth]{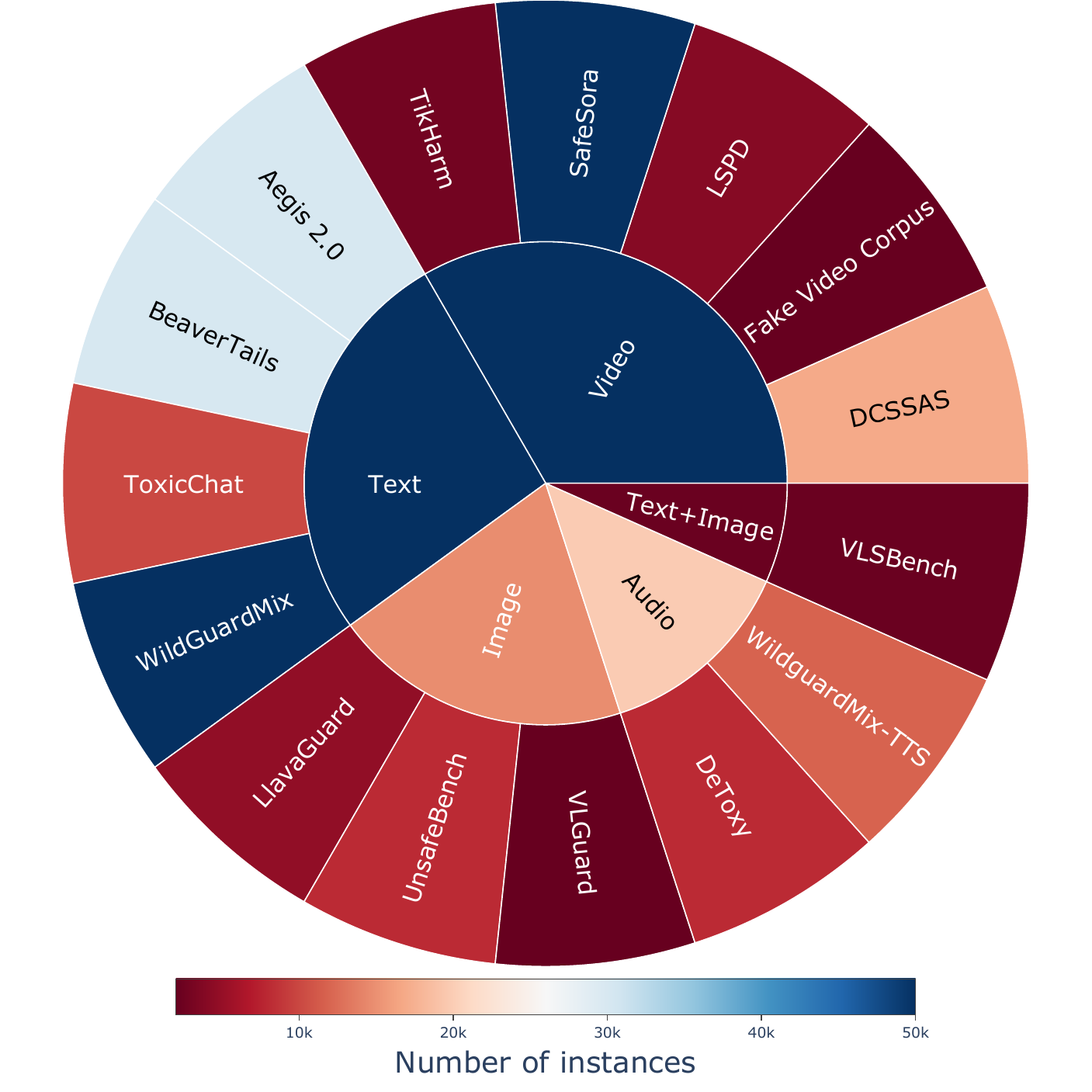}
    \caption{Collected datasets and the distribution of the constructed dataset.}
    \label{fig:sunburst}
    \vspace{-5mm} 
\end{figure}

Given the lack of an existing unified omni-modal safety finetuning dataset, we construct a comprehensive large-scale omni-modal safety dataset through targeted distillation to support the training process.
To increase the diversity of \texttt{input}, we first collect and aggregate datasets from both unimodal and cross-modal settings, including text, image, video, audio, and text-image modalities.
Each sample is paired with a corresponding binary safety label and, if unsafe, one or more associated violation categories.
Subsequently, we employ a targeted distillation process to extract safety reasoning knowledge from large expert models.
Given an input instance $\mathbf{x} \in \mathcal{X}$ consisting of one or more modalities, along with its ground-truth safety label $y$ and violation categories $\mathbf{c}$, the expert model $f_T$ produces a detailed natural language critique explaining decision:
\( f_T(\mathbf{x}, \mathbf{y}, \mathbf{c}) =  \mathbf{e_T}  \),
The formatted prompt used for targeted distillation is shown in~\cref{fig:distill_prompt}.
The outputs distilled from the expert models, together with the input, are used to construct the dataset $\mathcal{D}$:
\( \{ (\mathbf{x_i}, \mathbf{y_i}, \mathbf{c_i}, \mathbf{e_i}) \}_{i=1}^{N}. \)
An overview of collected datasets and the data distribution is presented in~\cref{fig:sunburst}.
Further statistics are summarized in~\cref{tab:benchmarks_details}.

To enhance interpretability and enable reasoning-based safety alignment, we augment each sample in the previously collected corpus with critiques generated by high-capacity models.
Specifically, we employ gpt-oss-120b~\cite{DBLP:journals/corr/abs-2508-10925} for textual data, Qwen3-VL-235B-A22B-Instruct~\cite{qwen3technicalreport} for visual-related (image, video, and the text-image pairs) data, and Kimi-Audio-7B-Instruct\cite{kimi_audio_2024} for auditory data.
For each instance, the teacher model is provided with the original content, its corresponding safety label, and the violated categories (if any), and is instructed to generate a reasoning critique explaining the rationale behind the safety assessment.
The complete prompting template used for critique generation is illustrated in~\cref{fig:distill_prompt}.

\subsubsection{Instruction Tuning}

Based on the distilled dataset $\mathcal{D}$ obtained from the omni-modal targeted distillation stage, we perform mission-focused instruction tuning to specialize the model toward the safety moderation.  
Specifically, we adopt an omni-modal instruction fine-tuning framework to enhance \model's capability to classify and reason about safety risks across modalities.

In our omni-modal setting, our goal is to (1) handle diverse input modalities (text, image, video, and audio), and (2) follow safety-specific instructions constrained by the policy guidelines $\mathcal{G}$ to perform unified, policy-grounded safety reasoning.  
Therefore, we leverage the omni-modal instruction-following dataset $\mathcal{D}$ and  optimize the model using a standard next-token prediction loss, enabling it to produce accurate safety judgments and coherent reasoning across modalities.

\subsubsection{Training Objective.}
The student model learns from constructed dataset $\mathcal{D}$, which contains omni-modal safety information, by minimizing a joint objective:
\(
\mathcal{L}_{\text{total}} =
\mathcal{L}_{\text{cls}} +
\mathcal{L}_{\text{cat}} +
\mathcal{L}_{\text{critique}},
\)
where $\mathcal{L}_{\text{cls}}$ is the classification loss for binary safety prediction, training the model to accurately discriminate between safe and unsafe content;
$\mathcal{L}_{\text{cat}}$ is the multi-label classification loss over violation categories, teaching the model to recognize and categorize fine-grained safety violations; and
$\mathcal{L}_{\text{critique}}$ is the autoregressive generation loss that aligns the student's critique with the teacher's explanation, enabling the model to produce interpretable critiques explaining the rationale behind the judgment.  
This training process transfers the policy alignment and safety reasoning capabilities from the teacher model to the guardrail model, allowing it to perform safety classification and justification in a unified manner across modalities.

\subsection{Reasoning-Based Inference}

Unlike simple classification-only guardrail models that output only a binary safety label, \model performs slow thinking inference to provide fine-grained and interpretable safety moderation. 
Specifically, it produces a structured output comprising the following components:

\begin{itemize}[itemsep=0em,leftmargin=1em]
    \item[(1)] \textbf{Safety judgment:} the overall safety assessment of the input, determining whether it is safe or unsafe.
    \item[(2)] \textbf{Violation categories:} the specific unsafe categories that the input violates, if the content is identified as unsafe.
    \item[(3)] \textbf{Reasoning critique:} a natural language explanation that articulates the rationale behind the model’s decision in accordance with the policy guidelines.
\end{itemize}
This formulation enables \model to go beyond shallow pattern recognition, supporting explainable analysis of potentially unsafe content across different modalities.

Formally, given a multimodal input $\mathbf{x} \in \mathcal{X}$ and a policy guideline set $\mathcal{G}$, \model computes:
\begin{equation}
g(\mathbf{x} \,|\, \mathcal{G}) = (\mathbf{\hat{y}}, \mathbf{\hat{c}}, \mathbf{\hat{e}}),
\end{equation}
where 
\( \mathbf{\hat{y}} \in \{\texttt{safe}, \texttt{unsafe}\}  \)
denotes the predicted safety label, $\mathbf{\hat{c}}$ represents the identified set of violation categories (empty if $\mathbf{\hat{y}} = \texttt{safe}$), and $\mathbf{\hat{e}}$ is the generated reasoning critique. 
The critique serves as an explicit intermediate representation of the model's decision process, offering insight into how the prediction aligns with the safety policy $\mathcal{G}$ and improving the transparency and interpretability of omni-modal safety moderation.

\section{Experiments}

\begin{table*}[t]
\centering
\small
\setlength{\tabcolsep}{5pt}{
\setlength{\aboverulesep}{0pt}
\setlength{\belowrulesep}{0pt}
\setlength{\extrarowheight}{.5ex}
\begin{tabular}{l|c|cc|cc|cc|cc|cc|>{\columncolor{gray!15}}c >{\columncolor{gray!15}}c}
    \toprule
    \multirow{2}{*}{\textbf{Model}}  
    & \multirow{2}{*}{\textbf{Size}}  
    & \multicolumn{2}{c|}{\textbf{BeaverTails}} 
    & \multicolumn{2}{c|}{\textbf{OpenAI}} 
    & \multicolumn{2}{c|}{\textbf{Toxic Chat}} 
    & \multicolumn{2}{c|}{\textbf{Aegis}} 
    & \multicolumn{2}{c|}{\textbf{WildGuard}} 
    & \multicolumn{2}{c}{\cellcolor{gray!15}\textbf{Average}}\\
    \cmidrule(lr){3-4} \cmidrule(lr){5-6} \cmidrule(lr){7-8} \cmidrule(lr){9-10} \cmidrule(lr){11-12} \cmidrule(lr){13-14}
    & & F1 & ACC & F1 & ACC & F1 & ACC & F1 & ACC & F1 & ACC & F1 & ACC \\
    \midrule
    GPT-4o & - & 83.5 & 72.6 & \textbf{82.3} & \textbf{88.6} & 51.2 & 94.8 & 54.3 & 65.6 & 78.0 & 89.1 & 69.9 & 82.1 \\
    Qwen3-235B & 235B & 81.9 & 77.1 & 80.1 & 85.4 & \underline{66.0} & \underline{95.5} & \underline{83.3} & \underline{83.6} & 74.4 & 90.2 & \underline{77.1} & 86.4 \\
    LLaMA-3.3-70B & 70B & 76.8 & 71.3 & 58.4 & 80.5 & 53.8 & 94.3 & 65.7 & 69.4 & 72.5 & 89.2 & 65.4 & 80.9 \\
    Qwen2.5-72B & 72B & \underline{83.6} & 80.5 & 79.8 & 85.2 & 54.4 & 94.3 & 68.9 & 73.2 & 72.2 & 89.1 & 71.8 & 84.5 \\
    Qwen2.5-Omni-7B & 7B & 58.4 & 55.4 & 70.0 & 70.8 & 65.2 & 93.9 & 77.3 & 73.9 & 62.0 & 83.1 & 66.6 & 75.4 \\
    Qwen2.5-7B & 7B & 75.3 & 72.6 & 72.6 & 81.4 & 58.3 & 95.1 & 75.4 & 77.5 & 63.3 & 86.2 & 69.0 & 82.6 \\
    LLaMA Guard 1 & 7B & 38.1 & 55.9 & 32.8 & 74.4 & 23.3 & 92.9 & 53.0 & 66.9 & 16.4 & 85.1 & 32.7 & 75.0 \\
    LLaMA Guard 2 & 8B & 72.3 & 73.5 & 74.4 & 85.6 & 30.9 & 93.7 & 59.0 & 69.2 & 68.9 & 89.9 & 61.1 & 82.4 \\
    LLaMA Guard 3 & 8B & 71.2 & 73.5 & \underline{81.6} & 85.7 & 37.7 & 93.3 & 65.8 & 73.5 & 73.5 & 91.6 & 66.0 & 83.5 \\
    ThinkGuard & 8B & 82.7 & \underline{81.6} & 78.7 & 79.0 & 49.8 & 92.8 & 69.9 & 74.6 & \underline{78.5} & \underline{92.5} & 71.9 & 84.1 \\
    \model-3B & 3B & 81.8 & \textbf{82.6} & 77.8 & 83.8 & \textbf{67.0} & \textbf{95.6} & 82.2 & 82.6 & 70.2 & 87.7 & 75.8 & \underline{86.5} \\
    \model-7B & 7B & \textbf{83.9} & 80.5 & 81.1 & \underline{87.9} & 58.2 & 95.3 & \textbf{84.0} & \textbf{84.1} & \textbf{78.6} & \textbf{92.4} & \textbf{77.2} & \textbf{88.0} \\
    \bottomrule
\end{tabular}
}
\caption{
Performance comparison of \model and baseline models on \textit{text-based safety benchmarks}.
\textbf{Bold} and \underline{underlined} values indicate the best and second-best performance, respectively.
}
\label{tab:text_benchmark_results}
\end{table*}
\begin{table*}[t]
    \centering
    \small
    \setlength{\tabcolsep}{5pt}{
\setlength{\aboverulesep}{0pt}
\setlength{\belowrulesep}{0pt}
\setlength{\extrarowheight}{.5ex}
    \begin{tabular}{l|c|cc|cc|cc|>{\columncolor{gray!15}}c >{\columncolor{gray!15}}c}
        \toprule
        \multirow{2}{*}{\textbf{Model}} & \multirow{2}{*}{\textbf{Size}} & \multicolumn{2}{c|}{\textbf{VLGuard}} & \multicolumn{2}{c|}{\textbf{UnsafeBench}} & \multicolumn{2}{c|}{\textbf{LlavaGuard}} & \multicolumn{2}{c}{\cellcolor{gray!15}\textbf{Average}} \\
        \cmidrule(lr){3-4} \cmidrule(lr){5-6} \cmidrule(lr){7-8} \cmidrule(lr){9-10}
         & & F1 & ACC & F1 & ACC & F1 & ACC & F1 & ACC \\
         \midrule
        GPT-4o & - & 75.5 & 79.7 & 55.2 & 74.7 & 68.3 & 74.6 & 66.3 & 76.3 \\
        Qwen3-VL-235B & 235B & 77.4 & 76.5 & \textbf{74.4} & \underline{80.9} & \underline{73.8} & 76.3 & \textbf{75.2} & 77.9 \\
        Qwen2.5-VL-72B & 72B & 78.5 & 77.4 & \underline{73.2} & 77.7 & 71.2 & 70.0 & 74.3 & 75.0 \\
        Qwen2.5-Omni-7B & 7B & 64.4 & 70.8 & 47.3 & 67.8 & 57.3 & 68.9 & 56.3 & 69.2 \\
        Qwen2.5-VL-7B & 7B & 62.8 & 48.2 & 55.7 & 40.1 & 59.0 & 46.5 & 59.2 & 44.9 \\
        LlavaGuard-v1.2-7B & 7B & 69.8 & 74.0 & 63.4 & 77.0 & 79.6 & 82.0 & 70.9 & 77.7 \\
        LLaMA Guard 3V & 11B & 0.0 & 55.8 & 0.0 & 61.9 & 0.0 & 75.0 & 0.0 & 64.2 \\
        \model-7B & 7B & \underline{79.1} & \underline{81.7} & 72.2 & \textbf{81.1} & 73.5 & \underline{77.1} & \underline{74.9} & \underline{80.0} \\
        \model-3B & 3B & \textbf{79.3} & \textbf{81.9} & 72.3 & \textbf{81.1} & \textbf{73.9} & \textbf{78.2} & \textbf{75.2} & \textbf{80.4} \\
        \bottomrule
    \end{tabular}
    }
\caption{
Performance comparison of \model and baseline models on \textit{image-based safety benchmarks}.
Best in \textbf{bold} and second-best in \underline{underlined}.
}
\vspace{-2mm}
\label{tab:image_benchmark_results}
\end{table*}
\begin{table*}[t]
    \begin{minipage}[t]{0.48\textwidth}
        \centering
        \small
        \setlength{\aboverulesep}{0pt}
        \setlength{\belowrulesep}{0pt}
        \setlength{\extrarowheight}{.5ex}
        \begin{tabular}{l|c|cc|cc}
            \toprule
            \multirow{2}{*}{\textbf{Model}} 
            & \multirow{2}{*}{\textbf{Size}} 
            & \multicolumn{2}{c|}{\textbf{SafeWatch}} 
            & \multicolumn{2}{c}{\textbf{SafeSora}} \\
            \cmidrule(lr){3-4} \cmidrule(lr){5-6}
            & & F1 & ACC & F1 & ACC \\
             \midrule
            GPT-4o & -& 84.2 & 77.5 & 49.9 & 82.2 \\
            Qwen3-VL-235B & 235B& 79.5 & 71.8 & 65.9 & 84.4 \\
            Qwen2.5-VL-72B & 72B& 72.5 & 64.0 & 68.0 & 83.7 \\
            LLaVA-Video-72B & 72B& 78.2 & 70.7 & 36.9 & 80.0 \\
            Qwen2.5-Omni-7B & 7B& 68.6 & 76.0 & 64.3 & 71.3 \\
            Qwen2.5-VL-7B & 7B& 49.7 & 46.2 & 62.2 & 83.5 \\
            LLaVA-Video-7B & 7B& 47.2 & 44.9 & 4.5 & 75.1 \\
            \model-3B & 3B& \textbf{92.3} & \underline{82.0} & \underline{70.1} & \underline{85.9} \\
            \model-7B & 7B& \underline{90.9} & \textbf{85.7} & \textbf{71.8} & \textbf{86.1} \\
            \bottomrule
        \end{tabular}
        \caption{
        Performance comparison of \model and baseline models on \textit{video-based safety benchmarks}.
        Best in \textbf{bold} and second-best in \underline{underlined}.
        }
        \label{tab:video_results_clean}
    \end{minipage}
    \hfill
    \centering
    \small
        \setlength{\aboverulesep}{0pt}
        \setlength{\belowrulesep}{0pt}
        \setlength{\extrarowheight}{.6ex}
    \begin{minipage}[t]{0.48\textwidth}
        \centering
        \begin{tabular}{l|c|cc|cc}
            \toprule
            \multirow{2}{*}{\textbf{Model}} 
            & \multirow{2}{*}{\textbf{Size}} 
            & \multicolumn{2}{c|}{\textbf{MuTox}} 
            & \multicolumn{2}{c}{\shortstack{\rule{0pt}{2.5ex}\textbf{WildGuard-}\\\textbf{TTS}}} \\
            \cmidrule(lr){3-4} \cmidrule(lr){5-6}
              & & F1 & ACC & F1 & ACC \\
             \midrule
            GPT-4o & & 38.7 & 66.1 & 81.6 & 85.2 \\
            Qwen2-Audio & 7B& 26.9 & 42.4 & 27.7 & 56.7 \\
            Qwen-Audio & 8B& 28.0 & 18.9 & 59.3 & 57.7 \\
            Kimi-Audio & 7B& 37.5 & 68.3 & 77.4 & 75.6 \\
            Qwen2.5-Omni-7B & 7B& 30.8 & 36.2 & 78.8 & 78.5 \\
            \model-3B & 3B& \underline{41.8} & \underline{72.3} & \textbf{88.4} & \textbf{89.8} \\
            \model-7B &7B& \textbf{43.7} & \textbf{75.4} & \underline{87.8} & \underline{89.2} \\
            \bottomrule
        \end{tabular}
        \caption{
        Performance comparison of \model and baseline models on \textit{audio-based safety benchmarks}.
        Best in \textbf{bold} and second-best in \underline{underlined}.
        }
        \label{tab:audio_results_clean}
    \end{minipage}
\end{table*}

In this section, we present comprehensive experimental results of \model.
We evaluate its performance under both unimodal and cross-modal settings on 15 guardrail and jailbreak benchmarks spanning four modalities — text, image, video, and audio.
We further design experiments to answer two central research questions:
(1) Does reasoning-based safety alignment enhance the omni-modal guardrail model’s ability to perform safety moderation and handle safety-critical challenges across diverse modalities?
(2) Can safety knowledge learned from seen modalities transfer to unseen ones, demonstrating cross-modal generalization in safety understanding and moderation capability?

\subsection{Experiment Settings.}

\textbf{Benchmarks.}
We evaluate \model on a diverse suite of public safety benchmarks spanning both unimodal and cross-modal settings.
For the unimodal setting, we assess performance across four modalities — text, image, video, and audio.
For text, we use BeaverTails~\cite{DBLP:conf/nips/JiLDPZB0SW023}, ToxicChat~\cite{DBLP:conf/emnlp/LinWTWGWS23}, WildGuardMix~\cite{DBLP:conf/nips/HanREJL00D24}, Aegis2.0\cite{DBLP:conf/naacl/GhoshVSPRVP25}, and the OpenAI Moderation dataset~\cite{DBLP:conf/aaai/MarkovZANLAJW23}.
For image, we adopt UnsafeBench~\cite{DBLP:journals/corr/abs-2405-03486}, VLGuard~\cite{DBLP:conf/icml/ZongBYYH24}, and LlavaGuard~\cite{helff2025llavaguard}.
For video, we evaluate on SafeSora~\cite{DBLP:conf/nips/DaiCWYCJ024} and SafeWatch-Bench~\cite{DBLP:conf/iclr/ChenPPL25}.
For audio, we use MuTox English split~\cite{DBLP:conf/acl/Costa-jussaMADH24} and WildGuardMix-TTS, which is constructed by converting WildGuardMix~\cite{DBLP:conf/nips/HanREJL00D24} test samples into speech using a text-to-speech pipeline consistent with our dataset construction procedure.
For the cross-modal setting, we evaluate \model on three configurations: image-text, video-text, and audio-text, corresponding to MM-SafetyBench~\cite{DBLP:conf/eccv/LiuZGLYQ24}, Video-SafetyBench~\cite{DBLP:journals/corr/abs-2505-11842}, and AIAH~\cite{DBLP:conf/naacl/YangQSH25}, respectively.
Further statistics are summarized in~\cref{tab:benchmarks_details}.
We employ Accuracy (ACC) and F1 as the primary evaluation metrics to assess safeguarding performance
For benchmarks containing only unsafe samples, we only report accuracy as the evaluation metric.

\textbf{Baselines.}
We compare \model against a comprehensive suite of baselines across all modalities, encompassing LLMs, VLLMs, and audio LLMs.
For each modality, we include both large-scale and small-scale state-of-the-art models to evaluate their safeguarding capabilities.
We also compare \model with available specialized guardrail models, including LLM-based and VLM-based guardrail models.
Detailed baseline configurations are summarized in~\cref{tab:baselines}.

\textbf{Training.}
We train two variants of our model, \model-7B and \model-3B, based on Qwen2.5-Omni-7B and Qwen2.5-Omni-3B, respectively.
Both models are trained using full-parameter supervised fine-tuning (SFT) on our constructed dataset.
Training is conducted on 8×H100 GPUs using the SWIFT training platform~\cite{zhao2024swiftascalablelightweightinfrastructure}.
We employ the AdamW optimizer with a learning rate of $1\times10^{-4}$, a cosine learning rate scheduler, and a warmup ratio of 0.05.
Each model is trained for 3 epochs with a per-device batch size of 2 for training and 1 for evaluation, and gradients are accumulated over 4 steps.
The random seed is fixed to 42 for reproducibility.
\subsection{Results.}
\paragraph{Uni-Modality.}
We compare the performance of \model against state-of-the-art proprietary and open-source baselines across four uni-modal safety scenarios: text (\cref{tab:text_benchmark_results}), image (\cref{tab:image_benchmark_results}), video (\cref{tab:video_results_clean}), and audio (\cref{tab:audio_results_clean}). Both \model-7B and \model-3B consistently achieve leading results across all modalities.
While \model-7B consistenly achieves strongest overall performance across all modalities, the smaller variant, \model-3B also can achieve results comparable to or exceeding much larger models such as GPT-4o, Qwen3-235B, and Qwen3-VL-235B, highlighting the effectiveness of our omni-modal safety alignment strategy.

\paragraph{Text.}
As shown in~\cref{tab:text_benchmark_results}, \model-7B achieves the highest average F1 and accuracy on text safety benchmarks, surpassing both large-scale proprietary models such as GPT-4o and open-source baselines including Qwen3-235B and LLaMA3.3-70B. It also consistently outperforms smaller general-purpose models as well as dedicated guardrail systems.
Specifically, \model-7B attains an average F1 of 77.2\% and an accuracy of 88.0\%, outperforming all other compared models and improving the average F1 by more than 10\% compared to LLaMA Guard 3.
Notably, the lighter variant, \model-3B, achieves performance comparable to Qwen3-235B while using only a fraction of its parameters.

\paragraph{Images.}
In the image domain (\cref{tab:image_benchmark_results}), \model-7B and \model-3B also exhibit higher unsafe content detection performance compared to all other baselines.
\model-7B achieves an average F1 of 75.2 \% and accuracy of 80.4 \%, matching the F1 score of Qwen3-VL-235B while using far fewer parameters.
The improvement over previous image safeguards is substantial.
LLaMA-Guard-3V~\cite{chi2024llamaguard3vision}, which is only designed for safeguarding multimodal conversational content, failed to provide safety assessment for image-only harmfulness evaluation, classfifying all the samples as safe.
This demonstrates the narraw focus of existing guardrail systems.

\paragraph{Videos.}
For the video-safety benchmarks (Table \ref{tab:video_results_clean}), both \model-7B and \model-3B achieve state-of-the-art performance. 
The improvement is especially pronounced on the SafeWatch-Bench, where both models exceed 90 \% on F1 score.
These results highlight the significant progress of \model in safety reasoning within video domain.
\paragraph{Audio.}
In the audio domain (see Table \ref{tab:audio_results_clean}), \model-7B and \model-3B also achieve superior performance across both audio safety benchmarks.
Since audio guardrails remain largely unexplored, our model provides a strong solution to the field and demonstrates that reasoning-based safety training can also effectively generalize to the audio modality.

\begin{figure*}[t]
    \centering
    \includegraphics[width=\linewidth]{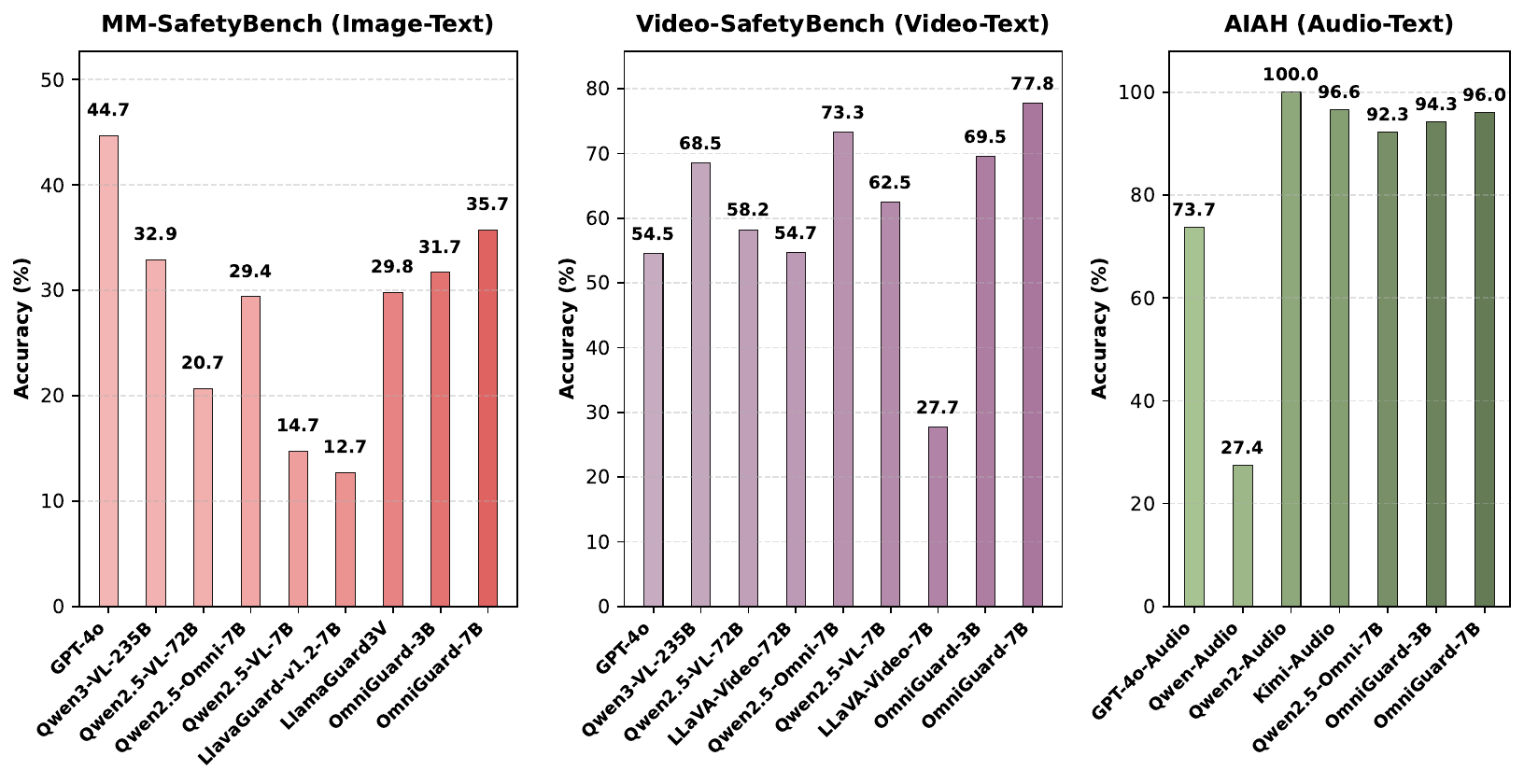}
    \vspace{-9mm} 
    \caption{
    Performance comparison of \model and baseline models on \textit{cross-modal safety benchmarks}.
    The performance is evaluated in Accuracy (ACC).
    }
    \label{fig:cross_modal}
    \vspace{-5mm} 
\end{figure*}

\paragraph{Cross-Modality.}
We further evaluate \model in cross-modal safety scenarios to assess its capability to reason across modalities.
As illustrated in Figure~\ref{fig:cross_modal}, our \model family demonstrates strong and consistent performance across all evaluated cross-modal safety benchmarks, including MM-SafetyBench, Video-SafetyBench, and AIAH.
Compared to Qwen2.5-Omni-7B, \model-7B consistently achieves significant improvements in accuracy across all benchmarks, highlighting the generalization of our omni-modal safety alignment in unifying multimodal reasoning.
Moreover, the lightweight \model-3B performs comparably to large-scale general-purpose models such as Qwen3-VL-235B on both MM-SafetyBench and Video-SafetyBench, despite having significantly fewer parameters.
These results further confirm that \model effectively generalizes safety reasoning across modalities, offering a scalable and parameter-efficient solution for cross-modal alignment.

\subsection{RQ1: Reasoning-Based Safety Training.}
\begin{figure*}[t]
    \centering
    \includegraphics[width=\linewidth]{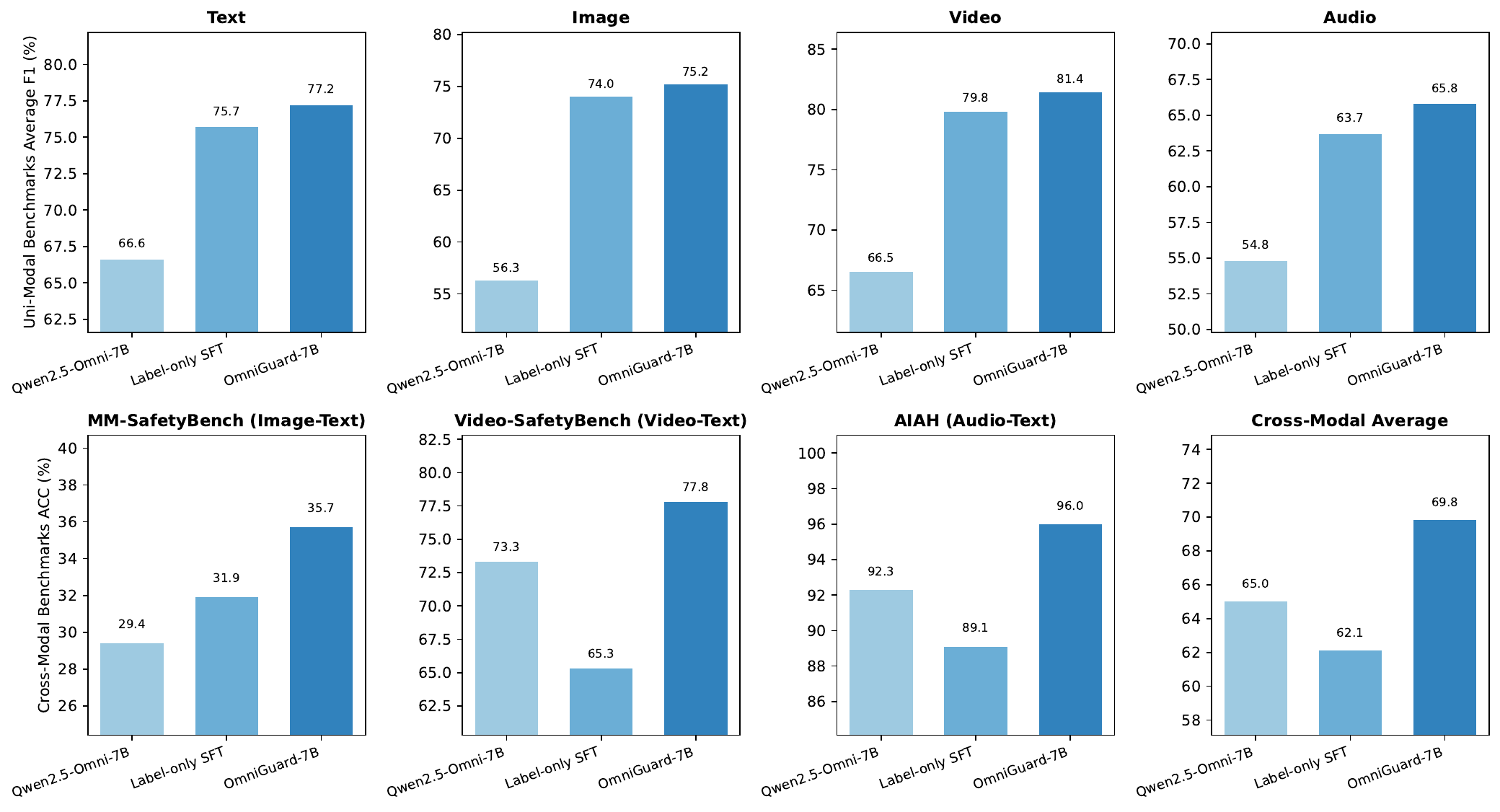}
    \vspace{-7mm}
    \caption{
Comparison of performance between Label-only SFT and critique-augmented training across both \textit{uni-modal} and \textit{cross-modal} settings. 
The upper four subplots show average performance results on uni-modal benchmarks (\textit{Text}, \textit{Image}, \textit{Video}, \textit{Audio}), evaluated by F1 score (\%).  
The bottom four subplots present cross-modal results on \textit{MM-SafetyBench (Image-Text)}, \textit{Video-SafetyBench (Video-Text)}, and \textit{AIAH (Audio-Text)}, along with the average performance, reported in accuracy (ACC, \%).  
    }
    \label{fig:reasoning}
\end{figure*}

To examine whether reasoning-based safety training enhances the performance of omni-modal guardrails and address the additional complexity introduced by omni-modal safety reasoning, we conduct further studies on the 7B model across four modalities, as shown in~\cref{fig:reasoning}.
We compare our \model-7B with the original base model Qwen2.5-Omni-7B~\cite{xu2025qwen25omnitechnicalreport} and its Label-only SFT variant, which is fine-tuned solely on safety classification labels without the curated reasoning traces used in our approach.

From these results, we draw several observations.  
(1) Both supervised fine-tuning methods can improve performance over the base model across all unimodal benchmarks (text, image, video, and audio), showing that simple safety fine-tuning can also enhance multimodal moderation capabilities.  
(2) Compared to the Label-only baseline, our reasoning-augmented training consistently achieves higher F1 scores across all unimodal settings, improving from 75.7→77.2 (text), 74.0→75.2 (image), 79.8→81.4 (video), and 63.7→65.8 (audio). This confirms that reasoning supervision helps the model better internalize safety assessment principles beyond surface-level pattern learning.  
(3) Notably, in the cross-modal setting, the Label-only SFT variant suffers a degradation in accuracy on the Video-SafetyBench and AIAH benchmarks, whereas our reasoning-augmented model achieves consistent gains across all three tasks. This suggests that simple label supervision fails to generalize effectively facing complex moderation tasks across modalities, while reasoning-based alignment endows the model with stronger guardrail understanding and transferability.

Overall, these results highlight that reasoning-based safety alignment not only enhances the performance of omni-modal guardrails across different modality settings, but also provides better understanding and generalization in complex cross-modal safety scenarios that label-only supervision fails to handle.


\subsection{RQ2: Cross-Modal Generalization.}

To investigate whether safety knowledge learned in seen modalities can generalize to unseen ones, we conduct cross-modal training and evaluation. Specifically, for each split, we train \model using data from three modalities for training and seen modality evaluation and leave one modality out for unseen modality evaluation.
We report the averaged F1 and accuracy across all four seen and unseen modality evaluations in Table~\ref{tab:cross_modal_results}.

We draw two main conclusions from these experiments.
(1) Overall, strong cross-modal transfer is observed across all four modalities.
The unseen modality performance remains close to seen modality results (79.4 vs. 81.8 F1 on average), indicating that \model successfully learns modality-invariant safety representations.
This suggests that harmful semantic patterns can be effectively aligned  and learned across text, image, video, and audio inputs.
\model further acquire generalizable safety reasoning ability across modalities.

(2) One notable exception arises in the audio modality, where the seen modality F1 (64.6\%) is slightly lower than the unseen modality F1 (65.1\%).
This phenomenon is due to the \model variant trained on text excluded split exhibiting degraded performance on the WildGuard-TTS benchmark.
WildGuardMix-TTS is the auditory version of WildGuardMix constructed by text-to-speech model.
Training without textual data also lead to degradation in the audio setting, this reveals that content which are semantically equivalent but are from different modalities (e.g., harmful text vs. its spoken version) can mutually influence each other during safety alignment.
And the knowledge from one form can be transferred to another semantically invariant form.

Taken together, these results demonstrate that cross-modal generalization in \model is substantial.
\model can generalizes safety reasoning from trained modalities to untrained ones and but also acquires modality-invariant semantic representations of unsafe content.

\begin{table}[h]
\centering
\small
\setlength{\tabcolsep}{9pt}{
\begin{tabular}{l|cc|cc}
\toprule
\multirow{2}{*}{\textbf{Modality}} & \multicolumn{2}{c|}{\textbf{Seen Modality}} & \multicolumn{2}{c}{\textbf{Unseen Modality}} \\
\cmidrule(lr){2-3} \cmidrule(lr){4-5}
 & \textbf{F1} & \textbf{ACC} & \textbf{F1} & \textbf{ACC} \\
\midrule
Text  & 88.1 & 80.1 & 84.7 & 77.4 \\
Image & 78.3 & 79.0 & 76.8 & 77.1 \\
Video & 78.9 & 85.1 & 76.6 & 80.0 \\
Audio & 64.6 & 92.2 & 65.1 & 90.2 \\
\midrule
Average & 81.8 & 84.1 & 79.4 & 81.2 \\
\bottomrule
\end{tabular}
}
\caption{
Performance comparison between \textit{seen modality} and \textit{unseen modality} settings across four modalities.
Metrics are F1 and Accuracy (\%). 
}
\label{tab:cross_modal_results}
\end{table}

\section{Conclusion}
In conclusion, we introduce \model-7B and \model-3B, the first family of omni-modal guardrails trained on a comprehensive and unified safety fine-tuning dataset covering both unimodal and cross-modal samples.
\model instantiates a unified omni-modal safety solution: it can moderate and reason about unsafe content in heterogeneous and cross-modal settings.
Extensive experiments demonstrate that \model-7B consistently outperforms existing guardrail models across all modalities, while \model-3B can achieves competitive results compared to large-scale LLMs and MLLMs such as Qwen3-235B and Qwen3-VL-235B.
These results highlight the strong omni-modal safety detection and reasoning capabilities of our approach, confirming the feasibility of a unified guardrail system with omni-modal understanding and interpretability.
Future work will investigate more complex and cross-modal safety scenarios to further advance omni-modal safeguarding in next-generation large language models.

\textbf{Limitations.}
Although \model demonstrates strong omni-modal safety reasoning and consistent performance across modalities, several limitations remain. 
While incorporating reasoning paths significantly enhances the interpretability and reliability of safety assessments, it inevitably increases inference latency due to the additional reasoning generation step.
As a result, \model has higher computational overhead compared to lightweight, binary-classification guardrail systems.
The trade-off between safety reasoning depth and inference latency can be further explored under different use cases and scenarios to further optimize safety robustness and efficiency.
Additionally, due to the limited availability of publicly accessible cross-modal safety fine-tuning datasets, there remains substantial room for progress in moderating more complex interleaved multimodal safety scenarios.
We hope future research will continue to improve the robustness and reliability of omni-modal systems, and advance their capability to safeguard against risks across complex modality combinations.

\bibliography{example_paper}
\bibliographystyle{icml2025}

\newpage
\appendix
\onecolumn

\section{Datasets and Benchmarks}
\begin{table*}[h!]
    \small
    \centering
    \begin{tabular}{lllrr}
        \toprule
         & Name & Citation & Train & Test\\
         \midrule
          \textbf{Text}
         & BeaverTails & \cite{DBLP:conf/nips/JiLDPZB0SW023} & 27,186 & 3,021\\
         & Aegis 2.0 & \cite{DBLP:conf/naacl/GhoshVSPRVP25} & 30,007 & 1,964\\
         & WildGuardMix & \cite{DBLP:conf/nips/HanREJL00D24} & 86,759 & 1,756\\
         & ToxicChat & \cite{DBLP:conf/emnlp/LinWTWGWS23} & 5,082 & 5,083\\
         & OpenAI Moderation & \cite{DBLP:conf/aaai/MarkovZANLAJW23} & – & 1,680\\
         \midrule
          
          \textbf{Image}
         & UnsafeBench & \cite{DBLP:journals/corr/abs-2405-03486} & 8,109 & 8,109\\
         & VLGuard & \cite{DBLP:conf/icml/ZongBYYH24} & 1,999 & 1,999\\
         & LlavaGuard & \cite{helff2025llavaguard} & 4,571 & 4,571\\
         \midrule
          
          \textbf{Video}
         & SafeSora & \cite{DBLP:conf/nips/DaiCWYCJ024} & 51,588 & 5,745\\
         & Fake Video Corpus & \cite{DBLP:journals/oir/PapadopoulouZPK19} & 380 & –\\
         & LSPD & \cite{phanlspd} & 4,000 & –\\
         & TikHarm & \cite{DBLP:conf/niles/BalatGBZ24} & 2,762 & –\\
         & DCSASS & \cite{DBLP:conf/cvpr/SultaniCS18} & 1,610 & –\\
         & SafeWatch-Bench & \cite{DBLP:conf/iclr/ChenPPL25} & – & 1620\\
         \midrule
          
          \textbf{Audio}
         & MuTox (English) & \cite{DBLP:conf/acl/Costa-jussaMADH24} & 13,617 & 1,945\\
         & WildGuardMix-TTS & \cite{DBLP:conf/nips/HanREJL00D24} & 10,000 & 1,756\\
         \midrule
          
          \textbf{Text-Image}
         & VLSBench & \cite{DBLP:conf/acl/HuLL0S25} & 2,240 & –\\
         & MM-SafetyBench & \cite{DBLP:conf/eccv/LiuZGLYQ24} & - & 5,040 \\         
         \midrule
          \textbf{Text-Video}
         & Video-SafetyBench & \cite{DBLP:journals/corr/abs-2505-11842} & – & 2,264\\
         \midrule
          \textbf{Text-Audio}
         & AIAH & \cite{DBLP:conf/naacl/YangQSH25} & – & 350\\
         \bottomrule
    \end{tabular}
    \caption{Overview of dataset sources used in the constructed dataset and benchmarks used for evaluation, with corresponding training and testing instance counts. “–” indicates not used or unavailable.}
    \label{tab:benchmarks_details}
\end{table*}



We next detail the data sources used in constructing dataset $\mathcal{D}$ and evaluation.
\vspace{-1em}
\paragraph{Text.} 
We collect and aggregate textual safety data from BeaverTails~\cite{DBLP:conf/nips/JiLDPZB0SW023}, WildGuardMix~\cite{DBLP:conf/nips/HanREJL00D24}, Aegis~2.0~\cite{DBLP:conf/naacl/GhoshVSPRVP25}, and ToxicChat~\cite{DBLP:conf/emnlp/LinWTWGWS23} for training and evaluation. Additionally, we include OpenAI Moderation~\cite{DBLP:conf/aaai/MarkovZANLAJW23} for evaluation.

\vspace{-1em}
\paragraph{Image.} 
We collect image safety data from UnsafeBench~\cite{DBLP:journals/corr/abs-2405-03486}, VLGuard~\cite{DBLP:conf/icml/ZongBYYH24}, and LlavaGuard~\cite{helff2025llavaguard} for both training and evaluation.

\vspace{-1em}
\paragraph{Video.}For constructing the dataset $\mathcal{D}$, we collect video safety data from SafeSora~\cite{DBLP:conf/nips/DaiCWYCJ024}, Fake Video Corpus~\cite{DBLP:journals/oir/PapadopoulouZPK19}, LSPD~\cite{phanlspd}, TikHarm~\cite{DBLP:conf/niles/BalatGBZ24}, and DCSASS~\cite{DBLP:conf/cvpr/SultaniCS18}.
From SafeSora, we utilize the generated video clips along with their corresponding safety classification labels.
Since the remaining datasets each target specific domains, we adopt the unified taxonomy proposed in \cite{DBLP:conf/iclr/ChenPPL25} to integrate them into a comprehensive video safety corpus. 
For evaluation, we include SafeSora~\cite{DBLP:conf/nips/DaiCWYCJ024}, and SafeWatch-Bench~\cite{DBLP:conf/iclr/ChenPPL25}.
\footnote{We did not include SafeWatch-Bench~\cite{DBLP:conf/iclr/ChenPPL25} for training, as their training split was unavailable at the time of our work.}

\vspace{-1em}
\paragraph{Audio.} 
For training and evaluation,
we leverage MuTox~\cite{DBLP:conf/acl/Costa-jussaMADH24}, a multilingual audio dataset for toxicity and harassment detection. 
We retain only the English subset. 
To further enhance the model’s capability in audio-based safety recognition, we also construct a synthetic dataset, WildGuardMix-TTS, by randomly sampling 10K textual prompts from WildGuardMix and converting them into audio clips using text-to-speech (TTS) model openaudio-s1-mini~\cite{fish-speech-v1.4}.

\vspace{-1em}
\paragraph{Cross-Modal.} Due to the scarcity of cross-modal safety training datasets with fine-grained annotations, we incorporate VLSBench~\cite{DBLP:conf/acl/HuLL0S25}, a carefully curated benchmark for visual-language safety evaluation, to strengthen the guardrail foundation for cross-modal safety alignment. For evaluation, we use MM-SafetyBench~\cite{DBLP:conf/eccv/LiuZGLYQ24}, Video-SafetyBench~\cite{DBLP:journals/corr/abs-2505-11842}, and AIAH~\cite{DBLP:conf/naacl/YangQSH25}, corresponding to image-text, video-text, and auio-text configurations.

\section{Formatted Prompt}

\begin{figure}[H]
\centering
\begin{prompt}{Formatted Prompt}
\footnotesize
\textbf{[Task]}
You are an LLM Judge tasked with evaluating the safety of input based on the Safety Guidelines. \\
Analyze the given input, identify any potential violations, and provide a detailed explanation for your judgment.\\

\textbf{\{Safety Guidelines\}}\\

\textbf{[Input]} \\
\{Text\} / <Image> / <Video> / <Audio> \\

\textbf{[Instructions]}\\
1. Identify whether the input is ``safe'' or ``unsafe''.\\
2. If ``unsafe'', list the violated categories separated by semicolons. \\
3. Provide a detailed explanation of whether the input is safe or unsafe. \\

\textbf{[Output Format]}\\
- \textbf{Safety Assessment:} \{Safety Label\}\\
- \textbf{Categories:} \{Violation Categories\}\\
- \textbf{Explanation:} [Provide detailed reasoning here]
\end{prompt}
\caption{Prompt template used for target distillation from teacher models when generating safety critiques. 
Provided with the safety label (safe or unsafe) and the corresponding violation categories, the teacher models are instructed to produce a detailed explanation describing the rationale behind the safety assessment.}
\label{fig:distill_prompt}
\end{figure}

\newpage
\section{Baselines}

\begin{table*}[h!]
\small
\centering
\begin{tabular}{llll}
\toprule
\textbf{Model} & \textbf{Citation} & \textbf{Size} & \textbf{Version}\\
\midrule
\multicolumn{4}{c}{\textbf{Large Language Models}} \\
\midrule
Qwen3-235B & \cite{qwen3technicalreport} & 235B & Qwen/Qwen3-235B-A22B-Instruct-2507 \\
LLaMA-3.3-70B & \cite{grattafiori2024llama3herdmodels} & 70B & Llama-3.3-70B-Instruct \\
Qwen2.5-72B & \cite{qwen2025qwen25technicalreport} & 72B & Qwen2.5-72B-Instruct \\
Qwen2.5-7B & \cite{qwen2025qwen25technicalreport} & 7B & Qwen2.5-7B-Instruct \\
LLaMA Guard 1 & \cite{DBLP:journals/corr/abs-2312-06674} & 7B & LlamaGuard-7b \\
LLaMA Guard 2 & \cite{metallamaguard2} & 8B & Meta-Llama-Guard-2-8B \\
LLaMA Guard 3 & \cite{grattafiori2024llama3herdmodels} & 8B & Llama-Guard-3-8B \\
ThinkGuard & \cite{wen-etal-2025-thinkguard} & 8B & ThinkGuard \\
\midrule
\multicolumn{4}{c}{\textbf{Vision Large Language Models}} \\
\midrule
Qwen3-VL-235B & \cite{qwen3technicalreport} & 235B & Qwen3-VL-235B-A22B-Instruct \\
Qwen2.5-VL-72B & \cite{bai2025qwen25vltechnicalreport} & 72B & Qwen2.5-VL-72B-Instruct \\
Qwen2.5-VL-7B & \cite{bai2025qwen25vltechnicalreport} & 7B & Qwen2.5-VL-7B-Instruct \\
LlavaGuard-v1.2-7B & \cite{helff2025llavaguard} & 7B & LlavaGuard-v1.2-7B-OV-hf \\
LLaMA Guard 3V & \cite{chi2024llamaguard3vision} & 11B & Llama-Guard-3-11B-Vision \\
LLaVA-Video-72B & \cite{zhang2025llavavideovideoinstructiontuning} & 72B & LLaVA-Video-72B-Qwen2 \\
LLaVA-Video-7B & \cite{zhang2025llavavideovideoinstructiontuning} & 7B & LLaVA-Video-7B-Qwen2 \\
\midrule
\multicolumn{4}{c}{\textbf{Audio Large Language Models}} \\
\midrule
Qwen2-Audio & \cite{chu2024qwen2audiotechnicalreport} & 7B & Qwen2-Audio-7B \\
Qwen-Audio & \cite{chu2023qwenaudioadvancinguniversalaudio} & 8B & Qwen-Audio-Chat \\
Kimi-Audio & \cite{kimiteam2025kimiaudiotechnicalreport} & 7B & Kimi-Audio-7B-Instruct \\
\midrule
\multicolumn{4}{c}{\textbf{Omni-Modal Large Language Models}} \\
\midrule
GPT-4o & \cite{openai2024gpt4ocard} & - & gpt-4o-2024-11-20,  gpt-4o-audio-preview-2025-06-03\\
Qwen2.5-Omni-7B & \cite{DBLP:journals/corr/abs-2503-20215} & 7B & Qwen2.5-Omni-7B \\
\bottomrule
\end{tabular}
\caption{Configuration details of baseline models used in evaluation, including Large Language Models (LLMs), Large Vision-Language Models (LVLMs), and Large Audio Language Models (LALMs).
“--” denotes information unavailable.
For GPT-4o, we employed gpt-4o-2024-11-20 for text, image, and video evaluations, and gpt-4o-audio-preview-2025-06-03 for audio-related assessments, as a truly omni-modal API endpoint was not publicly available at the time of evaluation.
}

\label{tab:baselines}
\end{table*}


\end{document}